\documentclass[runningheads]{llncs}

\usepackage{times}
\usepackage{soul}
\usepackage{url}
\usepackage{subfigure}
\usepackage{hyperref}

\usepackage[utf8]{inputenc}
\usepackage{graphicx}
\usepackage{amsmath}
\usepackage{booktabs}
\usepackage{multirow}
\usepackage{algorithm}
\usepackage{algorithmic}
\begin{document}

\title{One-Class Graph Neural Networks for Anomaly Detection in Attributed Networks}
%
%
\author{Xuhong Wang\inst{1} \and
Baihong Jin\inst{2} \and
Ying Du\inst{1} \and
Ping Cui\inst{1} \and
Yupu Yang\inst{1}} 
\authorrunning{X. Wang et al.}
%
\institute{Shanghai Jiao Tong University \\
\email{\{wang\_xuhong, duying, cuiping, ypyang\}@sjtu.edu.cn} \and
University of California, Berkeley\\\email{bjin@berkeley.edu}}

\maketitle              
\begin{abstract}
Nowadays, graph-structured data are increasingly used to model complex systems. Meanwhile, detecting anomalies from graph has become a vital research problem of pressing societal concerns. Anomaly detection is an unsupervised learning task of identifying rare data that differ from the majority. As one of the dominant anomaly detection algorithms, One-Class Support Vector Machine has been widely used to detect outliers. However, those traditional anomaly detection methods lost their effectiveness in graph data. Since traditional anomaly detection methods are stable, robust and easy to use, it is vitally important to generalize them to graph data. In this work, we propose One Class Graph Neural Network (OCGNN), a one-class classification framework for graph anomaly detection. OCGNN is designed to combine the powerful representation ability of Graph Neural Networks along with the classical one-class objective. Compared with other baselines, OCGNN achieves significant improvements in extensive experiments.

\keywords{Graph Anomaly Detection  \and Graph Neural Network \and Hypersphere Learning.}
\end{abstract}

\section{Introduction}
Nowadays, graph-structured data are increasingly used to model complex systems, ranging from social media networks~\cite{DBLP:conf/ijcai/PengLGSNLY19}, traffic networks~\cite{DBLP:conf/ijcai/YuYZ18} to financial nets~\cite{DBLP:conf/icdm/WangQL0JWFYZY19}. A graph is a structure amounting to a set of nodes $V$ in which some pairs of them are related by edges $E$. Through the introduction of edges between data instances, graph provides a powerful tool for effectively capture long-range correlations between them. Meanwhile, detecting anomalies from graph has become a vital research problem of pressing societal concerns~\cite{DBLP:journals/datamine/AkogluTK15}. Firstly, graph data with anomalies can disrupt the performance of machine learning algorithms and bring serious consequences. Secondly, graph anomaly detection (GAD) has significant applications in many security-related domains, e.g., discovering suspicious financial transactions, monitoring traffic jams, and unveiling malicious users in social networks.

Anomaly detection (or outlier detection) can be regarded as the task of identifying rare data items that differ from the majority of the data~\cite{DBLP:journals/sigpro/PimentelCCT14}. As one of the dominant anomaly detection algorithms, hypersphere learning approaches, e.g., Support Vector Data Description~(SVDD)~\cite{DBLP:journals/ml/TaxD04}, typically try to define a boundary around the given normal class data. Afterwards, whether the unknown data point is anomalous is determined by their location with respect to the boundary.

Although those traditional anomaly detection methods have been applied in diverse domains~\cite{DBLP:journals/sigpro/PimentelCCT14}, they may lose their effectiveness in graph data. Those methods can only handle Euclidean data (e.g., images, audio signal and texts). In the case of Euclidean data, data are treated as independent points lying in a multi-dimensional Euclidean space; however, nodes in graph may exhibit inter-dependencies, which means they are inherently related to one another. Graph embedding methods, such as DeepWalk~\cite{DBLP:conf/kdd/PerozziAS14}, can extract node relationship information as a fixed-length feature vector. Afterwards, traditional anomaly detection methods can be trained from that vector. However, these two-stage methods are decoupled in the sense that the graph embedding learning is task agnostic and not customized for GAD, which has an limited and indirect contribution to detecting anomalous nodes.

In addition to traditional anomaly detection techniques targeting Euclidean data, previous GAD works~\cite{DBLP:journals/datamine/AkogluTK15} have achieved empirical success. These works on GAD usually adopt graph matrix factorization to locate the outliers or find the target community at first and then detect outliers in a predefined subspace. However, several challenges exist in GAD tasks: (\romannumeral1) Data nonlinearity. The node feature and the interactions among nodes are often highly non-linear while many GAD methods rely on linear mechanisms. (\romannumeral2) Computational complexity. In the era of data deluge, real graphs can easily have millions of nodes and edges. The Computational overhead of classical GAD methods discourages their applicability to large-scale graphs. (\romannumeral3) Researchers and developers are proficient in simple off-the-shelf Euclidean anomaly detection techniques. Previous statistical graph feature based GAD methods increase their understanding and deployment costs, which greatly limits their development in the graph mining community. 

Graph Neural Networks (GNNs)~\cite{DBLP:conf/iclr/XuHLJ19} are a significant stride to operate directly on graph-structured data, and a promising method to address these above problems. GNNs are essentially a message passing (or neighborhood node aggregation) scheme, where each node aggregates feature information of its neighbors to compute its new feature vector. After multiple iterations of information aggregation, the feature vector of a node will capture the structural information among the node’s neighborhood. GNNs have non-linear activation functions and parallelization capabilities, which can address the data nonlinearity and the computational complexity issues, respectively. Even though GNNs have demonstrated outstanding performance in many graph mining tasks~\cite{DBLP:conf/ijcai/PengLGSNLY19,DBLP:conf/ijcai/YuYZ18,DBLP:conf/icdm/WangQL0JWFYZY19}, it remains unclear how to exploit their potentiality for GAD problem. 

To the best of our knowledge, little progress has been made to apply GNNs to GAD tasks. Ding~et~al.~\cite{DBLP:conf/sdm/DingLBL19} and Li~et~al.~\cite{DBLP:conf/cikm/LiHLDZ19} tried to solve the problems using graph autoencoder (GAE)~\cite{DBLP:journals/corr/KipfW16a}. They used graph convolutional network (GCN)~\cite{DBLP:conf/iclr/KipfW17}, a simple form of GNNs, to construct the encoder and decoder of AE, so that the node embedding can be extracted from the graph structure information as well as node attributes. However, GAE methods have the objectives of learning low-dimension node embedding and reconstructing the node attributes and connection relationship among nodes, which does not target GAD directly.

\begin{figure*}[t]
    \centering
    \includegraphics[width=0.9\textwidth]{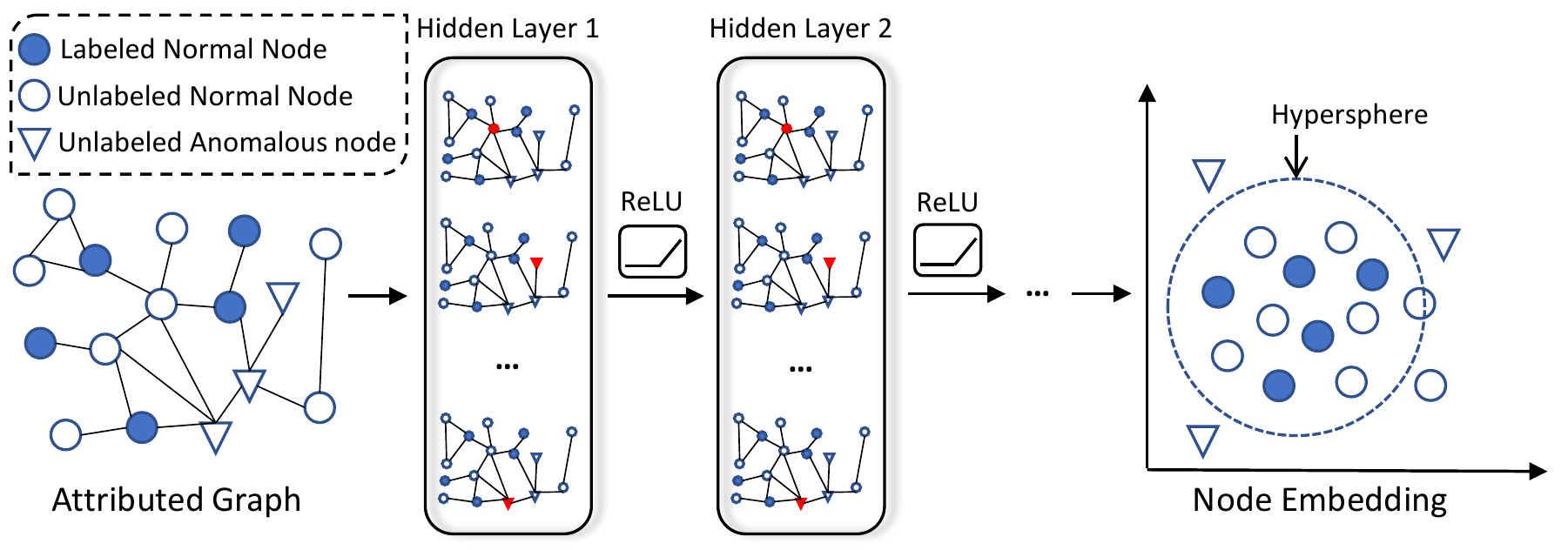}
    \caption{The overall framework of the proposed OCGNN. Guided by the proposed hypersphere learning objective, OCGNN learns to map the nodes into the embedding space and enclose the embeddings of normal nodes into a hypersphere centered at vector $c$ and with radius $r$. Afterwards, the nodes whose embeddings are located outside the hypersphere are regarded as anomalous nodes. Note that OCGNN is a hypersphere learning based GNN framework, so that people can use any forms of existing GNN layers in OCGNN.}
    \label{fig:overview}
\end{figure*}

There needs to be an end-to-end, easy-to-understand and powerful GAD technology. In this work, We propose One-Class Graph Neural Network (OCGNN), a hypersphere learning framework for GAD. As shown in Figure~\ref{fig:overview}, OCGNN is designed to combine the powerful representation ability of GNNs with the classical hypersphere learning objective to detect anomalies. Because the topology information on graphs is automatically extracted by the GNNs, users of OCGNN can handle GAD problems without deep understandings of graph theories. Furthermore, OCGNN is an end-to-end approach because the node representation learned by OCGNN is highly related to the GAD task, which means the embeddings learned from OCGNN is more friendly to the downstream anomaly detection tasks than those learned from DeepWalk. 
By aggregating the neighborhood information, an OCGNN computes the embedding of each observable node. Meanwhile, hypersphere learning objective aims at minimizing the volume of a hypersphere that encloses those embedding vectors~\cite{DBLP:conf/icml/RuffGDSVBMK18}. An anomalous node is defined by whether the location of its embedding is out of the hypersphere. 

Specifically, our contributions are listed as follows.
\begin{enumerate}
\item We propose a novel end-to-end OCGNN framework, which aims at applying the powerful representation ability of GNNs into graph anomaly detection tasks. To the best of our knowledge, this is one of the first GNN-based works in this scenario.
\item A hypersphere learning objective is proposed to drive the training of the GNN, thus OCGNN is a natural extension of One-Class Support Vector Machine (OCSVM) in the field of graph data. People can use OCGNN to handle GAD problem without any complicated graph theory, which will greatly facilitate the development of graph mining techniques and applications.
\end{enumerate}

\section{Related Works}
In this section, we briefly review related work in three aspects: (1) traditional anomaly detection, (2) graph neural networks and (3) graph based anomaly detection.

\subsection{Anomaly Detection in Euclidean Data}
Era of data deluge has grown the interest for anomaly detection techniques, especially for spotting anomalies in collections of multi-dimensional data points~\cite{DBLP:journals/sigpro/PimentelCCT14}. Owing to the prohibitive cost for accessing ground truth anomalies, most of existing methods are unsupervised, which can be divided into three main categories. 

(1) Distance-based approaches, such as Local Outlier Factor (LOF)~\cite{DBLP:conf/sigmod/BreunigKNS00} and Isolation Forest (IForest)~\cite{DBLP:conf/icdm/LiuTZ08}, assume that normal data are tightly clustered, while anomaly data are located far from their nearest neighbours.  (2) Boundary-based approach, e.g., SVDD~\cite{DBLP:journals/ml/TaxD04}, typically try to define a boundary around the normal class data. An unknown data that located outside the boundary is defined as an outlier. (3) Reconstruction-based approach assumes that anomalies cannot be effectively reconstructed from low-dimensional projections. In this category, PCA~\cite{Olive2017,baklouti2016iterated} and deep autoencoder models~\cite{DBLP:conf/kdd/ZhouP17,DBLP:conf/sdm/ChenSAT17,DBLP:journals/kbs/WangDLCSY20} are widely used, effective techniques to detect anomalies. Besides, Ruff~et~al.~\cite{DBLP:conf/icml/RuffGDSVBMK18} extended the shallow SVDD method to deep version, which shows promising results on image data. However, those three types of methods mentioned above are not applicable to the attributed graph.

\subsection{Graph Neural Networks}
Graph Neural Network (GNN) is a promising method to operate directly on non-Euclidean data. In GNN's neighborhood aggregation scheme, each node aggregates feature information of its neighbors to compute its new feature vector layer by layer. After multiple iterations of information aggregation, the feature vector of a node will captures the structural information among the node’s neighborhood. The embedding representation learned by GNNs can be used for downstream tasks such as node classification, link prediction, graph classification and so on.

According to the ways that information is aggregated, GNNs can be divided into two form: isotropic model and anisotropic model. Isotropic model assume that each neighbor contributes equally to the update of the central node. The typical isotropic GNNs are Graph Convolutional network (GCN)~\cite{DBLP:conf/iclr/KipfW17}, GraphSAGE~\cite{DBLP:conf/nips/HamiltonYL17} and Graph Isomorphism Network (GIN)~\cite{DBLP:conf/iclr/XuHLJ19}. On the other hand, anisotropic models assume that the importance of each neighbor to the central node is not equal, and use different weights to aggregate information from neighboring nodes through some mechanism (e.g., attention mechanism in Graph Attention Network (GAT)~\cite{DBLP:conf/iclr/VelickovicCCRLB18} or gating mechanism in Residual Gated Graph ConvNets (GatedGCN)~\cite{DBLP:journals/corr/abs-1711-07553}. 

Besides, graph Autoencoder (GAE)~\cite{DBLP:journals/corr/KipfW16a} is a natural extension of traditional AE to graph, which aim to learn low dimensional node vectors via a GCN built encoder and reconstruct the graph structure (adjacency matrix $\textbf{A}$) via a link prediction decoder. Nevertheless, all these GNN methods focus on learning embedded representations of nodes, and it remain unclear how to exploit the potential of GNNs for GAD.  

\subsection{Anomaly Detection on Attributed Graph}

Recently, we have witnessed an increasingly amount of efforts in spotting anomalies on attributed networks, and the existing methods can be divided into three categories: structure based, subspace based and residual based.

(1) Early GAD methods use graph structure information only such as community (common neighbors) analysis~\cite{DBLP:conf/kdd/GaoLFWSH10} and ego-net features~\cite{DBLP:conf/sdm/PerozziA16} to detect anomalies on the graph. (2) Besides that, researchers found that anomalies in attributed networks
could be spotted in only a subset of node attributes~\cite{DBLP:conf/icdm/SanchezMLKB13,DBLP:conf/kdd/PerozziASM14}. Although these subspace based methods can achieve superior results, they use little network structure information, which is undoubtedly reduces the learning efficiency of graph data. (3) Recently, residual analysis based methods~\cite{DBLP:conf/ijcai/LiDHL17,DBLP:conf/ijcai/PengLLLZ18} and GAE methods~\cite{DBLP:conf/sdm/DingLBL19,DBLP:conf/cikm/LiHLDZ19} leverage the graph structure information as well as its coherence with nodal attributes and have shown an increasing influence. Generally, the former methods use matrix decomposition characterizing and residual estimating techniques, but those shallow residual analysis mechanisms are not enough for today's computing demand for large-scale and non-linear data. On the other hand, GAE methods have the objectives of learning low-dimension node embedding and reconstructing the node attributes and connection relationship among nodes, which has an indirect contribution to anomaly detection on attributed graphs.

\section{One-Class Graph Neural Network}
As shown in Table~\ref{tab:notations}, we summarize the commonly used notations of this paper.

\subsection{Definitions}
In this section, we introduce notations and definitions of node-wise one-class classification and hypersphere learning. 

\textit{Definition 1.} \textbf{Attributed graph}: An attributed graph is $\mathcal{G}=(\mathbf{V}, \mathbf{E},\mathbf{X})$, where  $\mathbf{V}=\left\{v_{1}, \dots, v_{N}\right\}$ is the set of $N=|\mathbf{V}|$ nodes, $\mathbf{E} \subseteq \mathbf{V} \times \mathbf{V}$ is the set of $M=|\mathbf{E}|$ edges between nodes and $\mathbf{X} \in \mathcal{R}^{N \times D}$ denotes the features of the $N$ nodes. The graph structure can be represented by an adjacency matrix $\mathbf{A} \in \mathcal{R}^{N \times N}$, where $\mathbf{A}_{i,j}=1$ if there is an edge between node $v_i$ and $v_j$, otherwise $\mathbf{A}_{i,j}=0$.

\textit{Definition 2.} \textbf{Anomaly detection}: Given the anomaly-free training dataset $\{x_i , i =1, \ldots, K\}$, the model is trained to describe the boundary of normal data, and then model produces the anomaly score $S(x_u)$ for an unseen data point $x_u$. A data point with high anomaly score is defined as an outlier.

\textit{Definition 3.} \textbf{Node-wise graph anomaly detection}: Given a graph, all nodes $\mathbf{X}$, edges $\mathbf{A}$ and only part of the node labels are available for training. Afterwards, the labels of the remaining unlabeled nodes are predicted by the trained model. All nodes for training belong to one class (normal), while the remaining testing nodes belong to two classes (normal and abnormal). 


\begin{table}[t]
\centering
\begin{tabular}{ll}  
\toprule
Notations   & Descriptions \\
\midrule
$\mathbf{V}=\left\{v_{1}, \dots, v_{N}\right\}$       & The set of N nodes in a graph      \\
$\mathbf{V}_{tr}\subseteq\mathbf{V}$, $|\mathbf{V}_{tr}|=K$ & The set of K training nodes      \\
$\mathbf{X}\in \mathcal{R}^{N \times D}$ & Node feature matrix\\
$\mathbf{A}\in \mathcal{R}^{N \times N}$ & Adjacency matrix\\
$g\left(\mathbf{X},\mathbf{A} ; \mathcal{W}\right)$ & A graph neural network \\
$\mathbf{Z}\in \mathcal{R}^{N \times F}$  & Node embedding matrix, \\
$[\cdot]^{+}=\max (0, \cdot)$ & Non-negative operator     \\
$c\in \mathcal{R}^{F}$ & The hypersphere center  \\
$r\in \mathcal{R}^{+}$ & The hypersphere radius  \\
\bottomrule
\end{tabular}
\caption{Commonly Used Notations.}
\label{tab:notations}
\end{table}
\subsection{Hypersphere Learning}
Hypersphere learning is originally proposed in SVDD~\cite{DBLP:journals/ml/TaxD04}, which aims at learning an compact hypersphere boundary to cover all training data and detecting which (new) objects resemble this training set. In the application of anomaly detection, the training data are all normal, therefore hypersphere learning model can obtain a description boundary of normal data, then outliers can be precluded. 

Let $\mathcal{X} \subseteq \mathcal{R}^{d}$ be the data space, $\phi_{k}: \mathcal{X} \rightarrow \mathcal{F}_{k}$ be a mapping function from data space $\mathcal{X}$ to reproducing kernel Hilbert space $\mathcal{F}_{k}$, where $k: \mathcal{X} \times \mathcal{X} \rightarrow [0, \infty)$ is a positive-definite kernel function. The objective of SVDD is to describe the smallest hypersphere, in feature space $\mathcal{F}_{k}$, with center $c \in \mathcal{F}_{k}$ and radius $r>0$ that encloses the majority of the training data.
Given a training dataset $\mathcal{X}_{K}=\left\{x_{i} \in \mathcal{X}, i=1,\ldots,K\right\}$, SVDD solves the primal problem:
\begin{equation}
\begin{array}{c}{\min \limits _{r, c} r^{2}+\frac{1}{\beta K} \sum \limits_{i=1}^{K} \xi_{i}} \\ {\text { s.t. }\left\|\phi_{k}\left(x_{i}\right)-c\right\|_{\mathcal{F}_{k}}^{2} \leq r^{2}+\xi_{i}, \xi_{i} \geq 0,  \forall i}\end{array}
\label{eq:SVDD}
\end{equation}
where $\xi_{i}$ are non-negative slack variables to allow the contamination of outliers in the training dataset. The data points  $\phi_{k}\left(x_{i}\right) \in \mathcal{F}_{k}$ are not strictly inside the hypersphere, but the data located too away from the boundary should be penalized. The hyperparameter $\beta \in (0,1]$ controls the trade-off between the sphere volume and the penalties. After minimizing Eq.~\ref{eq:SVDD}, the center $c$ and the radius $r$ can be obtained, while $\xi_{i}$ is not a learnable parameter. Data point $x_{i}$ that falls outside the hypersphere, i.e., $\left\|\phi_{k}\left(x_{i}\right)-c\right\|_{\mathcal{F}_{k}}^{2}>r^{2}$, is defined as an outlier.

\subsection{Learning objective}
With classical hypersphere learning objective, an SVDD model builds a minimum volume hypersphere estimation of given data. However, our OCGNN learn useful node representations together with the hypersphere learning objective. To achieve this goal, we introduce GNN that is jointly trained to learning node embeddings by considering both node attributes and relationships and to keep the embeddings into a minimum hypersphere.

GNNs consider both the node attributes $\mathbf{X} \in \mathcal{R}^{N \times D}$ and the adjacency matrix $\mathbf{A} \in \mathcal{R}^{N \times N}$ (some GNN models use neighbor aggregation to replace the role of $\mathbf{A}$) when learning the node embedding vectors $\mathbf{Z} \in \mathcal{R}^{N \times F}$. Hence, We use $g(\mathbf{X},\mathbf{A};\mathcal{W})$ to represent a GNN with layer-wise trainable weights set $\mathcal{W}=\{\mathbf{W}^{(1)},\ldots,\mathbf{W}^{(L)}\}$ where $L \in \mathcal{N}$ is the number of hidden layers. For a specific $l^{th}$ layer, the forward propagation rule of GNN can be summarized as:
\begin{equation}
\mathbf{H}^{(l+1)}=g\left(\mathbf{H}^{(l)}, \mathbf{A} ; \mathbf{W}^{(l)}\right),
\label{eq:GNN}
\end{equation}
where $\mathbf{H}^{(l)}$ is the input for the $l^{th}$ GNN layer and $\mathbf{H}^{(l+1)}$ is the output after this layer. Note that the node attributes $\mathbf{X}$ is the input of the first layer, which equals to $\mathbf{H}^{(0)}$. The embedding matrix of the nodes $\mathbf{Z}$ is the final output $\mathbf{H}^{(L)}$. Thanks to the powerful GNN node embedding, it helps the performance improvement of down-stream tasks, such as node-level and graph-level classification and link prediction.

The aim of OCGNN is to jointly learn the net parameters $\mathcal{W}$ and minimize the volume of the data description hypersphere which is characterized by a radius $r \in \mathcal{R}^+$ and a center $c \in \mathcal{R}^{F}$. Given a graph defined by $(\mathbf{X},\mathbf{A})$ and the set of $K$ training nodes $\mathbf{V}_{tr}\subseteq \mathbf{V}$, where $K=\left|\left\{i: v_{i} \in \mathbf{V}_{tr}\right\}\right|$, the objective of OCGNN is set as:
\begin{equation}
\begin{aligned}
\mathcal{L}(r, \mathcal{W}) =\frac{1}{\beta K} \sum \limits_{v_i\in \mathbf{V}_{tr}} [\left\|g\left(\mathbf{X},\mathbf{A} ; \mathcal{W}\right)_{v_i}-c\right\|^{2}-r^{2}]^+ +r^{2}+\frac{\lambda}{2} \sum \limits_{l=1}^{L}\left\|\boldsymbol{W}^{(l)}\right\|^{2}
\label{eq:OCGNN}
\end{aligned}
\end{equation}

In the forward propagation, OCGNN receives all nodes and edges information in a graph and then output a node embedding matrix $\mathbf{Z}=g\left(\mathbf{X},\mathbf{A} ; \mathcal{W}\right), \mathbf{Z}\in \mathcal{R}^{N \times F}$. However, only $K$ node embeddings $\{\mathbf{Z}_{v_i}, v_{i}\in \mathbf{V}_{tr}\}$ are used in the loss function computation. 

The first term of Eq.\ref{eq:OCGNN} is a penalty for node embeddings lying out of the hypersphere, i.e., if the distance between an embedding vector and the center $c$ is greater than the radius $r$. The hyperparameter $\beta \in (0,1]$ controls the trade-off between the sphere volume and the penalties (we will discuss it later). As in classical SVDD, the second term, minimizing $r^2$, is to minimize the volume of sphere. The last term is the weight decay regularizer on the OCGNN network parameters $\mathcal{W}$ with a hyperparameter $\lambda >0$. 

The OCGNN objective lets the network learn to map the node embeddings that are closed the center $c$ of the sphere. Since the training nodes are all normal, OCGNN will extract the common factors of the given nodes. As a result, the description boundary of normal nodes can be obtained and the anomalous nodes can be detected.

For a node $v_i$ in the given graph, its anomaly score $S(v_i)$ can be defined by the location of the embedding respect to the sphere: $S(v_i)=\left\|g\left(\mathbf{X},\mathbf{A} ; \mathcal{W^{*}}\right)_{v_i}-c\right\|^{2}-r^{*2}$. If $S(v_i)>0$, the node $v_i$ is anomalous, otherwise it is a normal node. Note that the network parameters $\mathcal{W^{*}}$ and the learned radius $r^{*}$ can characterize a trained OCGNN model. The memory complexity of OCGNN is very low because no data need be stored for model prediction.

To clearly explain the role of $\beta$ in the training process, we define $\mathbf{\xi}=\frac{\lambda}{2} \sum \limits_{l=1}^{L}\left\|\boldsymbol{W}^{(l)}\right\|^{2}$ and $d_{v_i}=\left\|g\left(\mathbf{X},\mathbf{A}; \mathcal{W}\right)-c\right\|^{2}_{v_i}$ where $i=1, \dots, K$. Assuming that there is a node subset $\mathbf{V}_{o} \subset \mathbf{V}_{tr}$ whose nodes are mapped outside of the sphere by OCGNN, consisting $K_{o}=\left|\left\{i: d_{v_i}>r^{2}, v_{i}\in \mathbf{V}_{tr}\right\}\right|$ nodes. Thus the OCGNN optimization problem for $r$ can be written as
\begin{equation}
\begin{array}{lll}{} & {\underset{r}{\operatorname{argmin}}} & {r^{2}+\frac{1}{\beta K} \sum \limits_{v_i\in \mathbf{V}_{o}} (d_{v_i}-r^{2})+\mathbf{\xi}} 
\\ {=} & {\underset{r}{\operatorname{argmin}}} & {\left(1-\frac{K_{o}}{\beta K}\right)r^{2}+\frac{1}{\beta K} \sum \limits_{v_i\in \mathbf{V}_{o}} d_{v_i}+ \mathbf{\xi}}\\
{\Rightarrow} & {\underset{r}{\operatorname{argmin}}} & {\left(1-\frac{K_{o}}{\beta K}\right)r^{2}}
\end{array}
\label{eq:nu}
\end{equation}

Since the second and third term of Eq.\ref{eq:nu} is non-negative, radius $r$ is decreased as long as $K_o \leq \beta K$. That is, $\frac{K_o}{K} \leq \beta$ must satisfy in the optimum, which means $\beta$ is an upper bound on the fraction of anomalous training nodes. The optimal radius $r^*$ is determined by the largest $K_o$ that satisfies the inequality, because if the inequality is not satisfied, the model cannot continue to minimize $r$. Hence we also have $\frac{K_o+1}{K} > \beta$. $\beta$ is a lower bound on the fraction of samples being outside of the hypersphere. $\beta$ allows some nodes to be mapped out of the sphere, otherwise, in order to enclose all training nodes, the radius $r$ will be so large that OCGNN will fail to detect outliers. 

\subsection{The Paradigms of OCGNN} 
Our OCGNN is a GNN-based GAD framework, therefore OCGNN can be formed by any suitable GNN layer, such as Graph Convolutional network (GCN)~\cite{DBLP:conf/iclr/KipfW17}, Graph Attention Network (GAT)~\cite{DBLP:conf/iclr/VelickovicCCRLB18} and GraphSAGE~\cite{DBLP:conf/nips/HamiltonYL17}. In this section, we use GCN as an example to illustrate how OCGNN learns the node representation. If other forms of GNN layers are considered, people just need to replace the Eq.~\ref{eq:GCN} by other neighbor nodes aggregation function. GCN is one of the best paradigms of graph learning model. It is a multi-layer GNN model which expands the Eq.~\ref{eq:GNN} as follows: 
\begin{equation}
g\left(\mathbf{H}^{(l)}, \mathbf{A} ; \mathbf{W}^{(l)}\right)=\sigma\left(\tilde{\mathbf{D}}^{-\frac{1}{2}} \tilde{\mathbf{A}} \tilde{\mathbf{D}}^{-\frac{1}{2}} \mathbf{H}^{(l)} \mathbf{W}^{(l)}\right),
\label{eq:GCN}
\end{equation}
where $\tilde{\mathbf{A}}=\mathbf{A}+\mathbf{I}_N$ is the adjacency matrix of a graph whose nodes are added self-loop connections. $\tilde{\mathbf{D}}$ denotes the diagonal matrix of $\tilde{\mathbf{A}}$, where $\tilde{\mathbf{D}}_{i i}=\sum_{j} \tilde{\mathbf{A}}_{i j}$. $\tilde{\mathbf{D}}^{-\frac{1}{2}} \tilde{\mathbf{A}} \tilde{\mathbf{D}}^{-\frac{1}{2}}$ is the symmetric normalized adjacency matrix $\hat{\mathbf{A}}$, so we can directly compute $\hat{\mathbf{A}}$ as a pre-processing step. $\sigma(\cdot)$ is a non-linear activation function, such as ReLU. Note that the weight matrix $\mathbf{W}^{(l)}$ is shared for all nodes on the given graph. From the view of spectral graph theory~\cite{DBLP:conf/nips/DefferrardBV16}, GCN learns the node embedding $\mathbf{Z}_i$ by aggregating its own features $\mathbf{X}_i$ and first-order neighbor's features $\mathbf{X}_j$, where $j \in N(v_i)$, but the information of $k^{th}$-order neighborhood can be captured by stacking multiple GCN layers.


\subsection{Optimization}
\begin{algorithm}[tb]
\caption{Training OCGNN model}
\label{alg:OCGNN}
\begin{flushleft}
\textbf{Input}: Attributed Graph $\mathcal{G}=(\mathbf{V}, \mathbf{E},\mathbf{X})$, normal nodes set $\mathbf{V}_{tr}$, Slack parameter $\beta \in (0,1]$, weight dacay $\lambda >0$\\
\textbf{Output}: Weights $\mathcal{W}$, center $c \in \mathcal{R}^F$ and radius $r\in \mathcal{R}^+$
\end{flushleft}
\begin{algorithmic}[1] 
\STATE Initialize $\mathcal{W}$ using Glorot uniform initialization.
\STATE Initialize $r=0$, $c=\frac{1}{K}\sum \limits_{v_i \in \mathbf{V}_{tr}} g\left(\mathbf{X},\mathbf{A} ; \mathcal{W}\right)_{v_i}$.
\WHILE{epoch $<$ max epoch budget}
\STATE $\mathbf{d}_{\mathbf{V}_{tr}}=\left\|g\left(\mathbf{X},\mathbf{A} ; \mathcal{W}\right)_{\mathbf{V}_{tr}}-c\right\|^{2}$, $\mathbf{d}_{\mathbf{V}_{tr}} \in \mathcal{R}^K$
\STATE $\mathcal{L}=\frac{1}{\beta K} \sum \limits_{\mathbf{V}_{tr}} [\mathbf{d}_{\mathbf{V}_{tr}}-r^{2}]^{+}+r^{2}+\frac{\lambda}{2} \sum \limits_{l=1}^{L}\left\|\boldsymbol{W}^{(l)}\right\|^{2}$
\STATE Update $\mathcal{W}$ by its stochastic gradient $\nabla_{\mathcal{W}}\left(\mathcal{L}\right)$
\IF {epoch mod $\phi$ = 0}
\STATE Update $r$ using $(1-\beta) \times 100\%$ percentile of $\mathbf{d}_{\mathbf{V}_{tr}}$
\ENDIF
\ENDWHILE
\STATE \textbf{return} $\mathcal{W}$, $c$ and $r$
\end{algorithmic}
\end{algorithm}

The optimization process of OCGNN is summarized in Algorithm~\ref{alg:OCGNN}. A trained OCGNN model can be characterized by three parameters: weight matrix $\mathcal{W}$, radius $r$ and data center $c$. Like other neural network models, we use stochastic gradient descent to optimize the parameters $\mathcal{W}$ of the GNN model with OCGNN objective by back propagation (BP). Since radius $r$ is not a inner parameter of OCGNN network, $r$ and $\mathcal{W}$ can not be optimized synchronously by BP algorithm. Instead, we update the $r$ and $\mathcal{W}$ alternately during the training phase. First, we train $\mathcal{W}$ for $\phi \in \mathcal{N}$ epochs while $r$ is fixed. After every $\phi^{th}$ epochs, $r$ can be solved by simple linear percentile search. That is, for the training node set $\mathbf{V}_{tr}$, we can obtain a distance set $\mathbf{d}_{\mathbf{V}_{tr}}\in \mathcal{R}^K$. Afterward, we can sort $\mathbf{d}_{\mathbf{V}_{tr}}$ from small to large, and the radius $r$ can be defined via $(1-\beta)$ percentile of $\mathbf{d}_{\mathbf{V}_{tr}}$. Empirically, OCGNN is not sensitive to this parameter $\phi$. During the training phase, $c\in \mathcal{R}^{F}$ is fixed as $c_0$, which is computed from the mean of the training nodes embeddings by an initial forward propagation. To map the nodes around a target data center, $c_0$ is a good choice because most nodes are not too far away from $c_0$ in the node embedding space. 


\section{Experiments}
In this section, we introduce the detailed experimental setup and results, including datasets, baselines, network structures, hyperparameter selections, and performance analysis. The source code of our models is implemented using PyTorch and Deep Graph Library~\cite{wang2019dgl}. which will be released at a Github repository\footnote{\url{https://github.com/WangXuhongCN/myGNN}} as well as all baseline models for reproducibility.

\subsection{Datasets}
\begin{table}[t]
\centering
\begin{tabular}{@{}lrrrc@{}}
\toprule
\textbf{Datasets} & \textbf{Nodes} & \textbf{Edges} & \textbf{Features} & \textbf{Train/Val/Test nodes} \\ \midrule
\textbf{Cora}     & 2708           & 5429           & 1433           & 490/246/410                   \\
\textbf{Citeseer} & 3327           & 4732           & 3703           & 420/210/352                   \\
\textbf{Pubmed}   & 19717          & 44338          & 500            & 4725/2364/3936                \\ \bottomrule
\end{tabular}
\caption{Summary of the datasets used in our experiments.}
\label{tab:datasets}
\end{table}
Cora, Citeseer and Pubmed \cite{DBLP:journals/aim/SenNBGGE08} are publicly available and broadly used citation network datasets in previous studies~\cite{DBLP:conf/iclr/KipfW17,DBLP:conf/nips/HamiltonYL17,DBLP:conf/iclr/VelickovicCCRLB18}.  Scientific publications and their citation relationship are represented as nodes and edges on a graph, respectively. The node feature of each publication is described by a sparse bag-of-words feature vector computed from a dictionary. 
Our node anomaly detection datasets used in this paper is generated from these three plain node classification datasets by regarding one class as normal and the rest as anomalous. The normal classes in Cora, Citeseer and Pubmed datasets are ``Neural Networks'', ``IR'' and ``Diabetes Mellitus Type 2'', respectively. All the nodes in the train set pertain to the normal class, while, for the validation set or the test set, half of the nodes are normal and the other half are randomly sampled from the  anomalous class. The characteristics of all three datasets are summarized in Table~\ref{tab:datasets}, and our Github repository provides more details about our GAD datasets. For all the methods we use in this paper, the hyperparameters are tuned using the validation set, then the performances are evaluated in the test set.  
\subsection{Baselines}

Following~\cite{DBLP:conf/iclr/KipfW17,DBLP:conf/iclr/VelickovicCCRLB18,DBLP:conf/nips/HamiltonYL17}, we compared our OCGNN with twelve two-stage GAD methods and three state-of-the-art (SOTA) GAE based GAD methods. Two-stage methods are concatenated by the graph embedding methods and shallow anomaly detection methods. In the first stage, graph embedding is used to learn and map the graph structure information into a fixed length embedding vector. For the second stage, shallow Euclidean anomaly detection methods can be trained by the embedding vector and node feature vector. We use one of the most popular graph embedding methods, DeepWalk~\cite{DBLP:conf/kdd/PerozziAS14}, to learn a 128-dimensional embedding as the first stage method with the same hyperparameter reported in~\cite{DBLP:conf/iclr/KipfW17}. Four popular anomaly detection methods are used as the second stage and tuned by the performance in validation set: (\romannumeral1) OCSVM is our source method which we have described above. It has an important parameter $\beta$, which is chosen to be the same as in our OCGNN model. (\romannumeral2) Isolation Forest (IForest)~\cite{DBLP:conf/icdm/LiuTZ08} is an efficient method of detecting anomaly data in high-dimensional datasets. IF recursively and randomly splits the data feature and stores this in a forest data structure. Outliers are defined as the data who have shorter path in the tree because it means the data need a smaller number of splittings from other majority data.
(\romannumeral3) PCA~\cite{Olive2017} and AE, two popular reconstruction-based approaches, assume that anomalies are incompressible and thus cannot be effectively reconstructed from low-dimensional projections. 

Three SOTA GAE based GAD methods are Dominant (Dom)~\cite{DBLP:conf/sdm/DingLBL19}, a plain GAE~\cite{DBLP:journals/corr/KipfW16a} and an ablation model named GCN-AE. Dom is a recently proposed SOTA unsupervised GAD method on attribute graph. We reproduced Dom using the hyperparameter recommended by the authors and added the early stop technology. All these three models choose GCN layers to construct the encoder part of GAE, and the main difference between these three is the decoder structure as well as the training objective. Plain GAE proposed a inner product decoder to predict the unseen links on the graph, which equals to reconstruct the connection relationship between nodes. Dom designs an additional GCN-based decoder aiming at reconstruct the node attributes, so Dom has two different training objectives, reconstructing features and links, and a controlling parameter (set to be 0.5 as authors recommended) between them. To demonstrate how the two objectives affect the performance of GAE methods, we designed an ablation model named GCN-AE by removing the link prediction objective from Dom. 
\subsection{OCGNN Setup}
\begin{table*}[t]
\centering
\begin{tabular}{@{}clccc@{}}
\toprule
\multicolumn{2}{c}{Method}                        & Cora       & Citeseer   & Pubmed     \\ \midrule
\multirow{4}{*}{Raw Features}      & IForest      & 53.09 $\pm$ 0.03 & 46.33 $\pm$ 0.03 & 65.57 $\pm$ 0.02 \\
                                   & OCSVM        & 54.35 $\pm$ 0.02 & 57.05 $\pm$ 0.03 & 45.50 $\pm$ 0.01 \\
                                   & PCA          & 62.17 $\pm$ 0.01 & 58.10 $\pm$ 0.03 & 71.06 $\pm$ 0.01 \\
                                   & AE           & 62.17 $\pm$ 0.01 & 58.11 $\pm$ 0.03 & 71.05 $\pm$ 0.01 \\ \midrule
\multirow{4}{*}{DeepWalk}          & IForest      & 57.87 $\pm$ 0.02 & 51.00 $\pm$ 0.03 & 60.73 $\pm$ 0.01 \\
                                   & OCSVM        & 52.10 $\pm$ 0.03 & 43.13 $\pm$ 0.02 & 60.22 $\pm$ 0.01 \\
                                   & PCA          & 55.90 $\pm$ 0.03 & 46.65 $\pm$ 0.02 & 61.66 $\pm$ 0.01 \\
                                   & AE           & 55.91 $\pm$ 0.03 & 46.42 $\pm$ 0.02 & 61.66 $\pm$ 0.01 \\ \midrule
\multirow{4}{*}{DeepWalk+Raw Feat.}& IForest      & 53.56 $\pm$ 0.04 & 45.55 $\pm$ 0.06 & 65.60 $\pm$ 0.02 \\
                                   & OCSVM        & 51.59 $\pm$ 0.03 & 42.95 $\pm$ 0.02 & 60.10 $\pm$ 0.01 \\
                                   & PCA          & 62.38 $\pm$ 0.02 & 57.96 $\pm$ 0.03 & 72.04 $\pm$ 0.01 \\
                                   & AE           & 62.39 $\pm$ 0.02 & 57.96 $\pm$ 0.03 & 71.91 $\pm$ 0.01 \\ \midrule

\multirow{3}{*}{GAE based}      & GCN-AE       & 80.53 $\pm$ 0.05 & 59.52 $\pm$ 0.09 & 58.26 $\pm$ 0.02 \\
                                   & GAE~\cite{DBLP:journals/corr/KipfW16a}          & 60.15 $\pm$ 0.08 & 51.80 $\pm$ 0.03 & 54.27 $\pm$ 0.02 \\
                                   & Dom~\cite{DBLP:conf/sdm/DingLBL19}         & 67.50 $\pm$ 0.25 & 62.44 $\pm$ 0.15 & 53.92 $\pm$ 0.04 \\  \midrule                                 
\multirow{3}{*}{\textbf{Our OCGNNs}}              & \textbf{OC-GCN}       & 73.25 $\pm$ 0.02       &  62.81 $\pm$ 0.01          &  54.53 $\pm$ 0.01         \\
                                   & \textbf{OC-GAT}       &  \textbf{88.19} $\pm$ 0.02        &  79.06 $\pm$ 0.03          &  60.98 $\pm$ 0.01         \\
                                   & \textbf{OC-SAGE} & 86.97 $\pm$ 0.04          &  \textbf{85.62} $\pm$ 0.01      &  \textbf{74.72} $\pm$ 0.03         \\ \bottomrule

\end{tabular}
\caption{Average AUCs in \% with StdDevs (over 10 random seeds). Note that the best result is typeset in \textbf{bold}.}
\label{tab:results}
\end{table*}
Our OCGNN is a GNN framework for GAD, and it can be formed by any suitable GNN layer. To evaluation the performance of OCGNN, we use three popular GNN layer paradigms, GCN~\cite{DBLP:conf/iclr/KipfW17}, GAT~\cite{DBLP:conf/iclr/VelickovicCCRLB18} and GraphSAGE~\cite{DBLP:conf/nips/HamiltonYL17}, whose model configurations are almost same. To better demonstrate the true performance of the OCGNN framework, we use the same hyperparameters for training in all datasets and OCGNN paradigms. The penalty parameter $\beta$ is set as 0.1. The OCGNN model is initialized with Glorot uniform weight initialization~\cite{DBLP:journals/jmlr/GlorotB10} and optimized by the AdamW~\cite{DBLP:conf/iclr/LoshchilovH19} SGD optimizer with a learning rate of 0.001. During training, we apply weight decay regularization
with $\lambda= 0.0005$. We trained each OCGNN model using an early stopping strategy on both the OCGNN loss (Eq.~\ref{eq:OCGNN}) and AUC score on the validation set, with a maximum of 5000 epochs and a patience of 100 epochs. For Cora and Citeseer datasets, we apply a three-layers OCGNN with the hidden size of 64-64-32. For Pubmed dataset, we use a two-layers OCGNN with hidden size of 128-64. For each individual OCGNN model, the dimension of output layer (the dimension of node embedding) is always the half of the hidden size. After each GNN layer, the dropout layer of 0.5 rate and the ReLU activation function is applied. The aggregator type of each GraphSAGE layer is set as pooling, and the attention heads of GAT layer is set as 8. Note that the structure and hyperparameters used in this study proved to be sufficient for our applications, although they can still be improved. 
\begin{figure*}[t]
    \centering
    \subfigure[Raw features]
    {
        \label{fig:Rawdata}
        \includegraphics[width=0.3\textwidth]{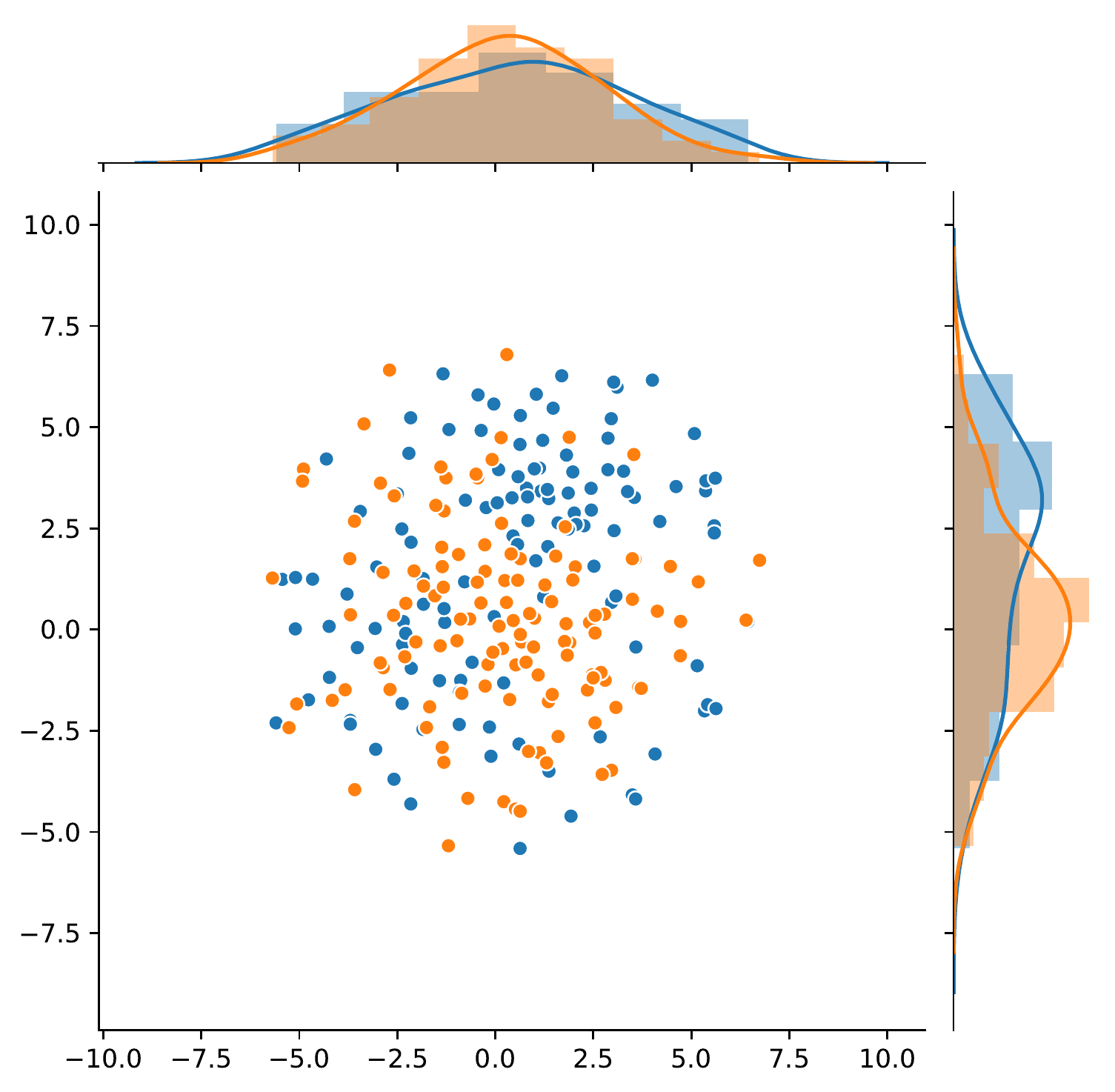}
    }
    \subfigure[Initialized model]
    {
        \label{fig:initialized}
        \includegraphics[width=0.3\textwidth]{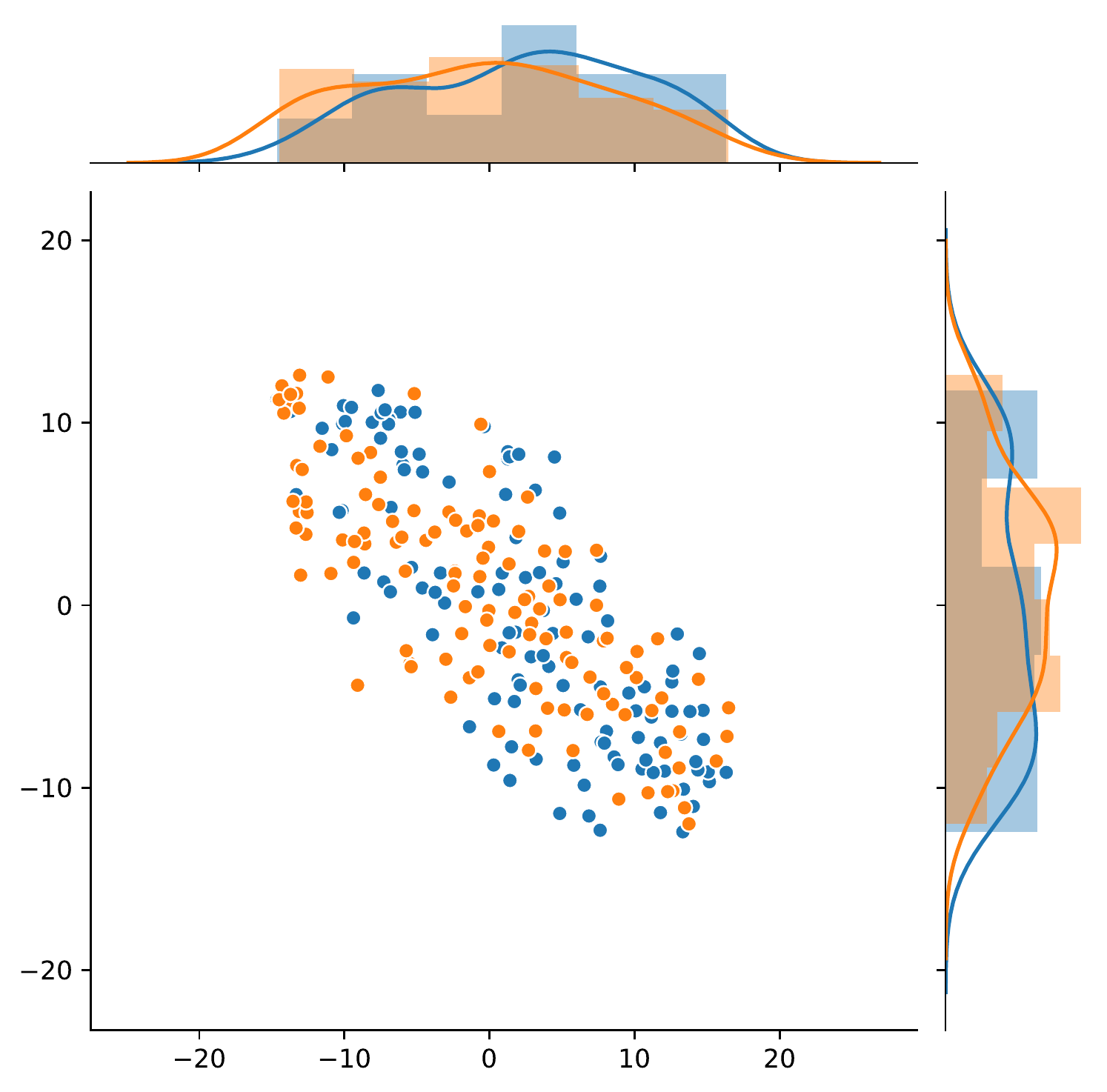}
    }
    
    \subfigure[After 500 epochs]
    {
        \label{fig:After500}
        \includegraphics[width=0.3\textwidth]{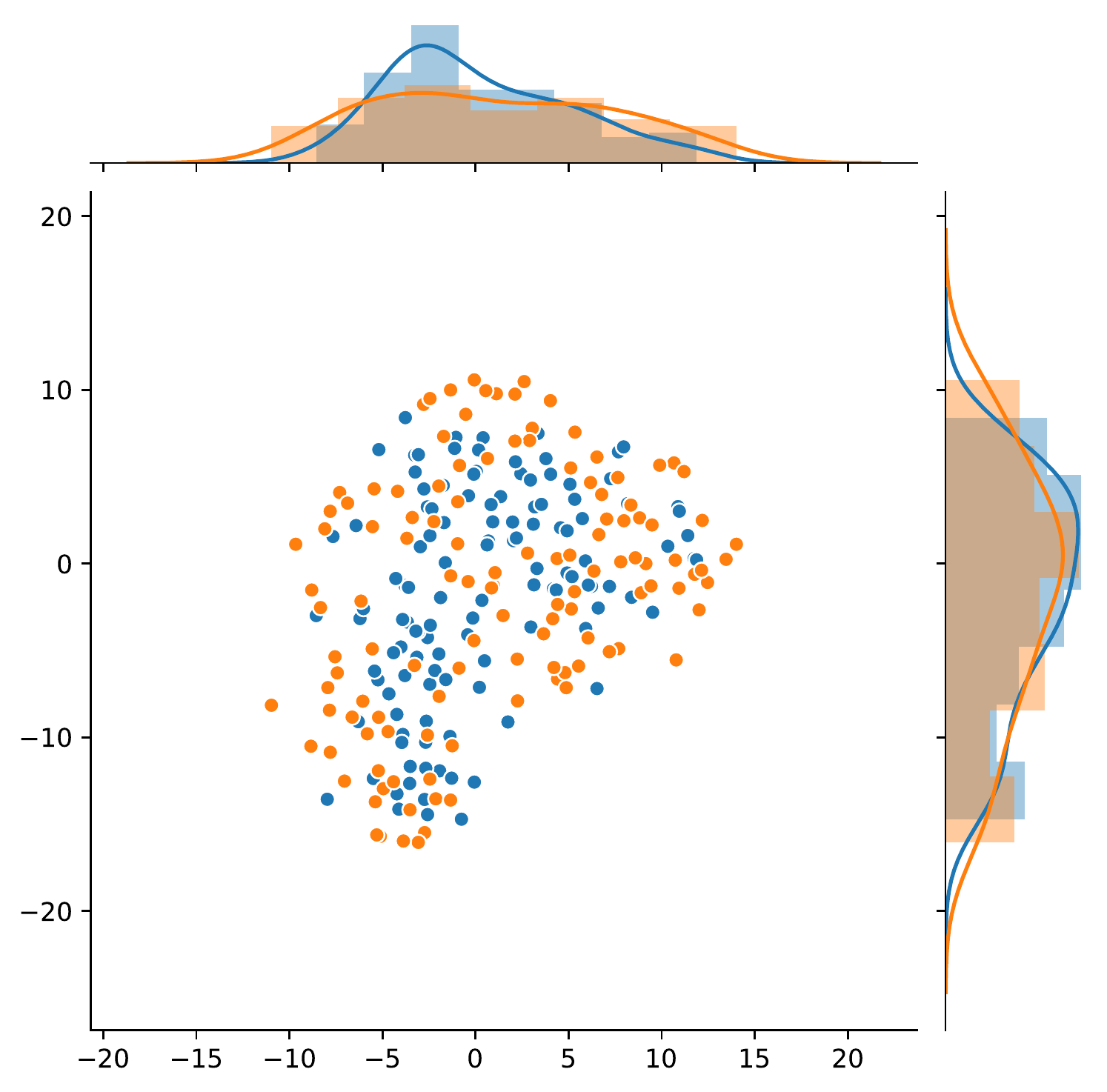}
    }
    \subfigure[Well-trained model]
    {
        \label{fig:trained}
        \includegraphics[width=0.3\textwidth]{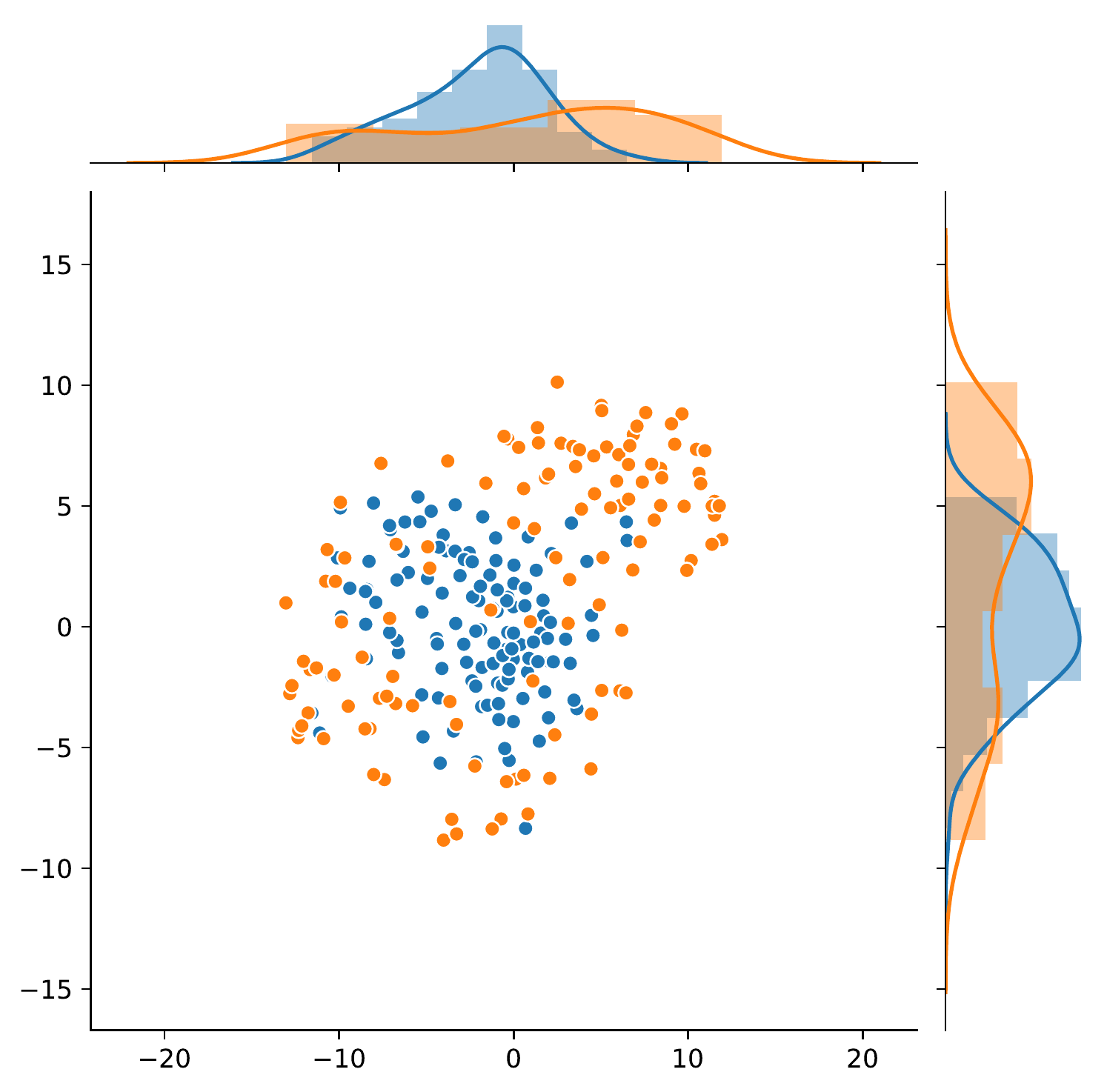}
    }
    \caption{T-SNE visualization of the node embeddings in Cora dataset. The KDE curves at top and right show the probability distribution in each dimension. Blue and orange points represent the normal nodes and the anomalous nodes, respectively. (a) The visualization of nodes' raw features. (b) The visualization of node embeddings from a randomly initialized OCGNN model. (c) Node embeddings after 500 epochs training. (d) Node embeddings from a learned OCGNN model. We can conclude that OCGNN does learn the features of normal nodes and can distinguish abnormal nodes in an unsupervised manner.
    }
    \label{fig:zspace}
\end{figure*}
\begin{figure*}[t]
    \centering
    \includegraphics[width=0.75\textwidth]{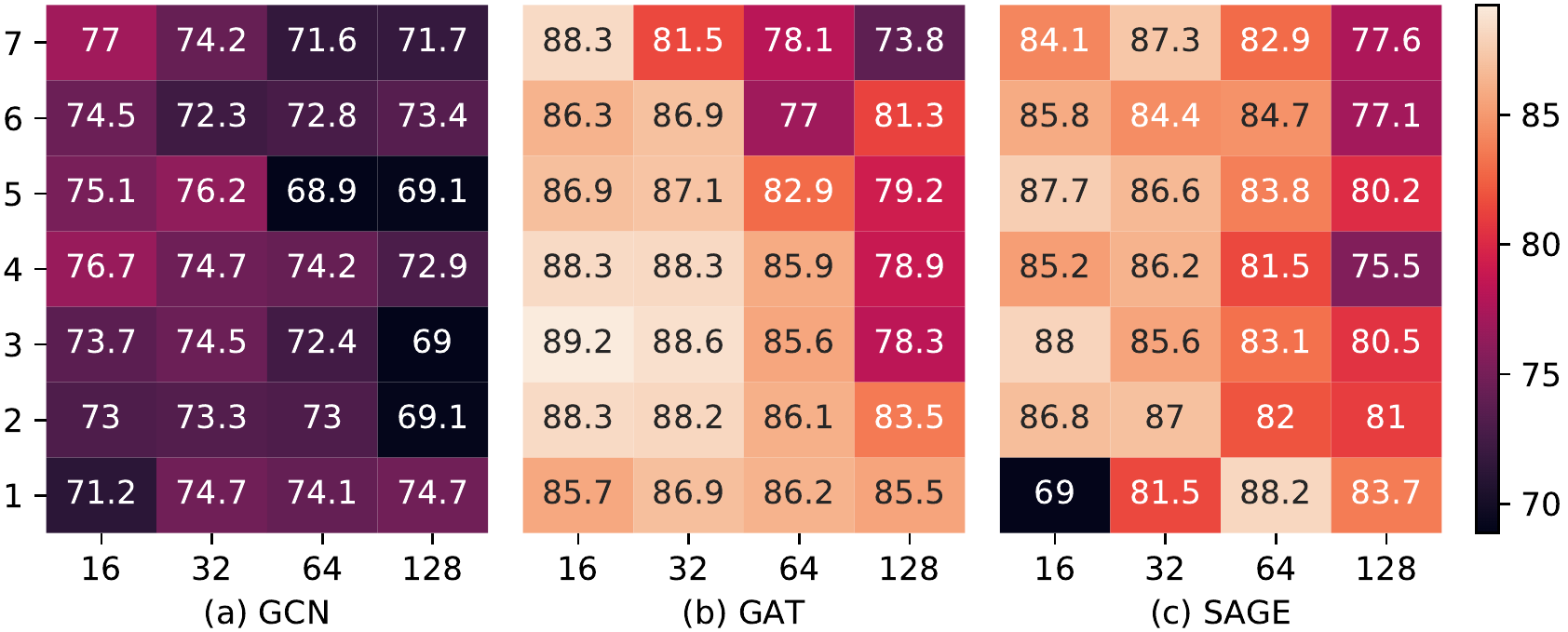}
    \caption{The heatmap of average AUCs (\%) of diverse OCGNN models in Cora dataset. The horizontal axis represents the dimensions of the hidden layer, and the vertical axis represents the number of layers in the network. The lighter the color, the higher the AUC performance of the model. Basically, the performance of the OCGNN model is robust to the structure of the network.}
    \label{fig:hidden-layer-heat}
\end{figure*}
\subsection{Results}

Table~\ref{tab:results} shows the experiment results at AUC metrics. According to the experiment results, our OCGNN model achieve best results in all datasets. Between three forms of OCGNNs, we believe that OC-SAGE is the most powerful and stable one especially in larger datasets like Pubmed, because OC-SAGE has great robustness because of its simple neighbor aggregation strategy. Since the complicated attention mechanism, OC-GAT perform well in the small Cora dataset, but lost superiority in Pubmed. There is no significant performance gap between OC-GCN and Dom, but OC-GCN has fewer net parameters and lower time complexity than Dom, which we will explain in next section.

Among the twelve the two-stage methods, only AE and PCA achieve competitive results in Pubmed dataset with raw features. The reason is that PCA and AE are better at processing high-dimensional original features. OCGNN outperforms the other two-stage methods, reaching the highest AUC improvement of above 30\% than OCSVM does in Cora and Citeseer dataset. Among three GAE based methods, we can conclude that GCN-AE (reconstructing node attributes) is more effective than GAE (reconstructing the adjacency matrix), because it is exceedingly hard to find the anomalous disturbance in such a super high dimensional adjacency matrix. Dom is an assembly model that has both the above two tasks. The loss value of the two objectives in Dom may not be on the same order of magnitude, which can cause instabilities during training.


We performed a T-SNE~\cite{maaten2008visualizing} visualization analyses on the embeddings learned by the OCGNN framework in order to better understand its properties. We focus our analysis exclusively on the Cora dataset and use the GraphSAGE algorithm to form our OCGNN. The reasons are that Cora dataset has the smallest number of nodes which can significantly aid clarity and GraphSAGE is a outstanding graph learning layer that is simple in form and has fewer hyperparameters.
The embedding visualization in Fig.~\ref{fig:Rawdata} and~\ref{fig:initialized} indicates that we cannot detect anomalous nodes by the raw node features or the initialized OCGNN. After 500 epochs of training, the normal and anomalous nodes seem to have a tendency to separate. The KDE curves in Fig.~\ref{fig:trained} prove that where the probability density of normal nodes is high, the density of anomalous nodes is low, and vice versa. It means that a majority of normal and anomalous nodes can be easily partitioned in the embedding space, if the model converges.

\subsection{Robustness}

To verify the impact of network structure on performance, we tune the number of hidden layers from 1 to 7 and adjust the number of hidden neurons in each hidden layer from 16 to 128, respectively. The results from different network structures are shown in Figure~\ref{fig:hidden-layer-heat} as heatmaps. OCGNN's performance will decrease when the size of the network is too large (top right) or too small (bottom left). The best results are usually obtained with a 2- to 4-layer model and 16 or 32 dimensions. When the network capacity is extremely small (1 hidden layer and 16 dimensions), the performance of OCGNN is greatly reduced because the insufficient network capacity cannot learn enough information from the normal training data. On the other hand, OCGNNs achieve slightly worse results when the network is too deep or too wide. A deep GNN model usually suffers from the over-smooth problem~\cite{DBLP:conf/icml/XuLTSKJ18} because GNN models iteratively aggregate the node information of almost the entire graph layer by layer, and calculate node embeddings of low diversity. If the network is too wide, curse of dimensionality will cause OCGNN to fail in calculating distance terms in the loss function, because the distances of data points are approaching the same in high-dimensional space. As long as the scale of neural network is large enough, the number of network layers or the number of neurons will not have a significant impact on the results.
\begin{figure*}[t]
    \centering
    \includegraphics[width=0.65\textwidth]{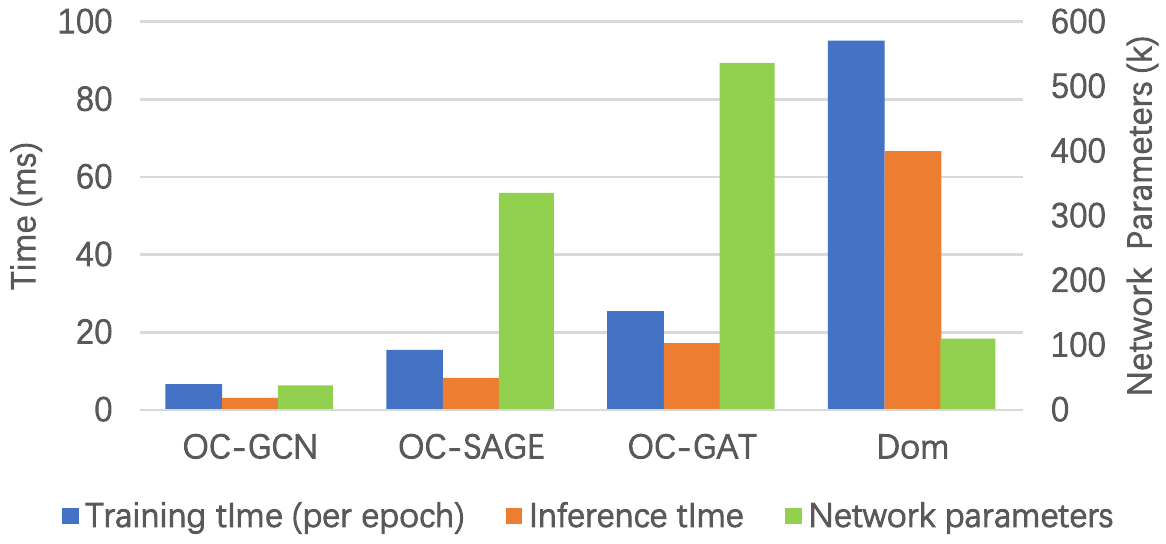}
    \caption{Runtime experiments on the pubmed dataset, It shows the training and test runtimes and number of network parameters for the different approaches. Dom has three times as many net parameters than that of OC-GCN, and 15 times slower than OC-GCN. Whereas, OC-SAGE and OC-GAT have above ten times as many parameters but do not induce much increase in runtime.}
    \label{fig:model-time}
\end{figure*}

\subsection{Runtime}
We conducted the runtime experiments on a Linux server with an Intel Xeon CPU and an NVIDIA GeForce GTX 1080Ti GPU. Figure~\ref{fig:model-time} summarizes the training and test runtimes and number of network parameters for the different approaches. GCN-based models have fewer network parameters but still result in worse AUC performance, as shown in Table~\ref{tab:results}. Dom has three times as many net parameters than OC-GCN because Dom is a three-networks model: an encoder, an attribute decoder and a structure decoder, while our OCGNN only has one encoder. This means that Dom has three times as many layers as the OCGNN model does, which will increase the gradient BP time. Besides that, the cost function of Dom is to compute the mean square error between high-dimensional input ($\mathbf{X}$ and $\mathbf{A}$) and their reconstruction matrix, whereas OCGNN's loss is a computation between a low-dimensional embedding matrix $\mathbf{Z}$ and a data center vector $c$. Therefore, the time complexity of Dom is significant higher than our OCGNN model. Although the network scales of OC-SAGE and OC-GAT are much more larger than OC-GCN, they only increase the time complexity by a few times, because these three OCGNN models have the same numbers of GNN layer and Dom have almost three times numbers of layer, which is the core factor of time complexity.

\section{Conclusion}
In this paper, we proposed One-Class Graph Neural Network (OCGNN), a one-class classification framework for graph anomaly detection. OCGNN aims at mapping the training nodes into a hypersphere in the embedding space, through the powerful representation ability of Graph Neural Networks. Our results from extensive experiments demonstrate that the proposed OCGNN achieves significant improvements. In the future, we will generalize OCGNN into graph-level and dynamic node anomaly detection. 

\section*{Acknowledgments}
This research is partially supported by *******.



\bibliographystyle{splncs04}
\bibliography{ocgnn}

\begin{thebibliography}{10}
\providecommand{\url}[1]{\texttt{#1}}
\providecommand{\urlprefix}{URL }
\providecommand{\doi}[1]{https://doi.org/#1}

\bibitem{DBLP:journals/datamine/AkogluTK15}
Akoglu, L., Tong, H., Koutra, D.: Graph based anomaly detection and
  description: a survey. Data Min. Knowl. Discov.  \textbf{29}(3),  626--688
  (2015)

\bibitem{baklouti2016iterated}
Baklouti, R., Mansouri, M., Nounou, M., Nounou, H., Hamida, A.B.: Iterated
  robust kernel fuzzy principal component analysis and application to fault
  detection. Journal of Computational Science  \textbf{15},  34--49 (2016)

\bibitem{DBLP:journals/corr/abs-1711-07553}
Bresson, X., Laurent, T.: Residual gated graph convnets. CoRR
  \textbf{abs/1711.07553} (2017)

\bibitem{DBLP:conf/sigmod/BreunigKNS00}
Breunig, M.M., Kriegel, H., Ng, R.T., Sander, J.: {LOF:} identifying
  density-based local outliers. In: ACM SIGMOD International Conference on
  Management of Data {(SIGMOD)}. pp. 93--104. {ACM} (2000)

\bibitem{DBLP:conf/sdm/ChenSAT17}
Chen, J., Sathe, S., Aggarwal, C.C., Turaga, D.S.: Outlier detection with
  autoencoder ensembles. In: {SDM}. pp. 90--98. {SIAM} (2017)

\bibitem{DBLP:conf/nips/DefferrardBV16}
Defferrard, M., Bresson, X., Vandergheynst, P.: Convolutional neural networks
  on graphs with fast localized spectral filtering. In: {NeurIPS}. pp.
  3837--3845 (2016)

\bibitem{DBLP:conf/sdm/DingLBL19}
Ding, K., Li, J., Bhanushali, R., Liu, H.: Deep anomaly detection on attributed
  networks. In: {SDM}. pp. 594--602. {SIAM} (2019)

\bibitem{DBLP:conf/kdd/GaoLFWSH10}
Gao, J., Liang, F., Fan, W., Wang, C., Sun, Y., Han, J.: On community outliers
  and their efficient detection in information networks. In: {KDD}. pp.
  813--822. {ACM} (2010)

\bibitem{DBLP:journals/jmlr/GlorotB10}
Glorot, X., Bengio, Y.: Understanding the difficulty of training deep
  feedforward neural networks. In: {AISTATS}. vol.~9, pp. 249--256. JMLR.org
  (2010)

\bibitem{DBLP:conf/nips/HamiltonYL17}
Hamilton, W.L., Ying, Z., Leskovec, J.: Inductive representation learning on
  large graphs. In: {NIPS}. pp. 1024--1034 (2017)

\bibitem{DBLP:journals/corr/KipfW16a}
Kipf, T.N., Welling, M.: Variational graph auto-encoders. CoRR
  \textbf{abs/1611.07308} (2016)

\bibitem{DBLP:conf/iclr/KipfW17}
Kipf, T.N., Welling, M.: Semi-supervised classification with graph
  convolutional networks. In: {ICLR} (Poster). OpenReview.net (2017)

\bibitem{DBLP:conf/ijcai/LiDHL17}
Li, J., Dani, H., Hu, X., Liu, H.: Radar: Residual analysis for anomaly
  detection in attributed networks. In: {IJCAI}. pp. 2152--2158. ijcai.org
  (2017)

\bibitem{DBLP:conf/cikm/LiHLDZ19}
Li, Y., Huang, X., Li, J., Du, M., Zou, N.: Specae: Spectral autoencoder for
  anomaly detection in attributed networks. In: {CIKM}. pp. 2233--2236. {ACM}
  (2019)

\bibitem{DBLP:conf/icdm/LiuTZ08}
Liu, F.T., Ting, K.M., Zhou, Z.: Isolation forest. In: {ICDM}. pp. 413--422.
  {IEEE} Computer Society (2008)

\bibitem{DBLP:conf/iclr/LoshchilovH19}
Loshchilov, I., Hutter, F.: Decoupled weight decay regularization. In: {ICLR}
  (2019)

\bibitem{maaten2008visualizing}
Maaten, L.v.d., Hinton, G.: Visualizing data using t-sne. Journal of machine
  learning research  \textbf{9}(Nov),  2579--2605 (2008)

\bibitem{Olive2017}
Olive, D.J.: Principal Component Analysis, pp. 189--217. Springer (2017)

\bibitem{DBLP:conf/ijcai/PengLGSNLY19}
Peng, H., Li, J., Gong, Q., Song, Y., Ning, Y., Lai, K., Yu, P.S.: Fine-grained
  event categorization with heterogeneous graph convolutional networks. In:
  {IJCAI}. pp. 3238--3245. ijcai.org (2019)

\bibitem{DBLP:conf/ijcai/PengLLLZ18}
Peng, Z., Luo, M., Li, J., Liu, H., Zheng, Q.: {ANOMALOUS:} {A} joint modeling
  approach for anomaly detection on attributed networks. In: {IJCAI}. pp.
  3513--3519. ijcai.org (2018)

\bibitem{DBLP:conf/sdm/PerozziA16}
Perozzi, B., Akoglu, L.: Scalable anomaly ranking of attributed neighborhoods.
  In: {SDM}. pp. 207--215. {SIAM} (2016)

\bibitem{DBLP:conf/kdd/PerozziASM14}
Perozzi, B., Akoglu, L., S{\'{a}}nchez, P.I., M{\"{u}}ller, E.: Focused
  clustering and outlier detection in large attributed graphs. In: {KDD}. pp.
  1346--1355. {ACM} (2014)

\bibitem{DBLP:conf/kdd/PerozziAS14}
Perozzi, B., Al{-}Rfou, R., Skiena, S.: Deepwalk: online learning of social
  representations. In: {KDD}. pp. 701--710. {ACM} (2014)

\bibitem{DBLP:journals/sigpro/PimentelCCT14}
Pimentel, M.A.F., Clifton, D.A., Clifton, L.A., Tarassenko, L.: A review of
  novelty detection. Signal Process.  \textbf{99},  215--249 (2014)

\bibitem{DBLP:conf/icml/RuffGDSVBMK18}
Ruff, L., G{\"{o}}rnitz, N., Deecke, L., Siddiqui, S.A., Vandermeulen, R.A.,
  Binder, A., M{\"{u}}ller, E., Kloft, M.: Deep one-class classification. In:
  {ICML}. Proceedings of Machine Learning Research, vol.~80, pp. 4390--4399.
  {PMLR} (2018)

\bibitem{DBLP:conf/icdm/SanchezMLKB13}
S{\'{a}}nchez, P.I., M{\"{u}}ller, E., Laforet, F., Keller, F., B{\"{o}}hm, K.:
  Statistical selection of congruent subspaces for mining attributed graphs.
  In: {ICDM}. pp. 647--656. {IEEE} Computer Society (2013)

\bibitem{DBLP:journals/aim/SenNBGGE08}
Sen, P., Namata, G., Bilgic, M., Getoor, L., Gallagher, B., Eliassi{-}Rad, T.:
  Collective classification in network data. {AI} Magazine  \textbf{29}(3),
  93--106 (2008)

\bibitem{DBLP:journals/ml/TaxD04}
Tax, D.M.J., Duin, R.P.W.: Support vector data description. Mach. Learn.
  \textbf{54}(1),  45--66 (2004)

\bibitem{DBLP:conf/iclr/VelickovicCCRLB18}
Velickovic, P., Cucurull, G., Casanova, A., Romero, A., Li{\`{o}}, P., Bengio,
  Y.: Graph attention networks. In: {ICLR} (Poster). OpenReview.net (2018)

\bibitem{DBLP:conf/icdm/WangQL0JWFYZY19}
Wang, D., Qi, Y., Lin, J., Cui, P., Jia, Q., Wang, Z., Fang, Y., Yu, Q., Zhou,
  J., Yang, S.: A semi-supervised graph attentive network for financial fraud
  detection. In: {ICDM}. pp. 598--607. {IEEE} (2019)

\bibitem{wang2019dgl}
Wang, M., Yu, L., Zheng, D., Gan, Q., Gai, Y., Ye, Z., Li, M., Zhou, J., Huang,
  Q., Ma, C., Huang, Z., Guo, Q., Zhang, H., Lin, H., Zhao, J., Li, J., Smola,
  A.J., Zhang, Z.: Deep graph library: Towards efficient and scalable deep
  learning on graphs. ICLR Workshop on Representation Learning on Graphs and
  Manifolds  (2019)

\bibitem{DBLP:journals/kbs/WangDLCSY20}
Wang, X., Du, Y., Lin, S., Cui, P., Shen, Y., Yang, Y.: advae: {A}
  self-adversarial variational autoencoder with gaussian anomaly prior
  knowledge for anomaly detection. Knowl. Based Syst.  \textbf{190},  105187
  (2020)

\bibitem{DBLP:conf/iclr/XuHLJ19}
Xu, K., Hu, W., Leskovec, J., Jegelka, S.: How powerful are graph neural
  networks? In: {ICLR}. OpenReview.net (2019)

\bibitem{DBLP:conf/icml/XuLTSKJ18}
Xu, K., Li, C., Tian, Y., Sonobe, T., Kawarabayashi, K., Jegelka, S.:
  Representation learning on graphs with jumping knowledge networks. In:
  {ICML}. Proceedings of Machine Learning Research, vol.~80, pp. 5449--5458.
  {PMLR} (2018)

\bibitem{DBLP:conf/ijcai/YuYZ18}
Yu, B., Yin, H., Zhu, Z.: Spatio-temporal graph convolutional networks: {A}
  deep learning framework for traffic forecasting. In: {IJCAI}. pp. 3634--3640.
  ijcai.org (2018)

\bibitem{DBLP:conf/kdd/ZhouP17}
Zhou, C., Paffenroth, R.C.: Anomaly detection with robust deep autoencoders.
  In: {KDD}. pp. 665--674. {ACM} (2017)

\end{thebibliography}

\end{document}